\documentclass[10pt,twocolumn,letterpaper]{article}

\usepackage[pagenumbers]{iccv} 

\definecolor{iccvblue}{rgb}{0.21,0.49,0.74}
\definecolor{tabblue}{rgb}{0.12,0.49,0.85}
\usepackage{hyperref}
\hypersetup{pagebackref,breaklinks,colorlinks,allcolors=iccvblue}
\usepackage{multirow}
\usepackage{pifont} 

\usepackage{algorithm}
\usepackage{algpseudocode}

\def\confName{ICCV}
\def\confYear{2025}
\begin{document}
\title{S3R-GS: Streamlining the Pipeline for Large-Scale Street Scene Reconstruction}

\author{
Guangting Zheng\textsuperscript{1}\thanks{Equal contribution to this work.}, 
Jiajun Deng\textsuperscript{2}\footnotemark[1], 
Xiaomeng Chu\textsuperscript{1},
Yu Yuan\textsuperscript{1}, 
Houqiang Li\textsuperscript{1}, Yanyong Zhang\textsuperscript{1} \\
\textsuperscript{1}University of Science and Technology of China \textsuperscript{2}The University of Adelaide\\
\tt\small\{zgt,cxmeng,yyhappier\}@mail.ustc.edu.com,jiajun.deng@adelaide.edu.au,\{lihq,yanyongz\}@ustc.edu.cn\\
}

\maketitle

\begin{abstract}
Recently, 3D Gaussian Splatting (3DGS) has reshaped the field of photorealistic 3D reconstruction, achieving impressive rendering quality and speed. 
However, when applied to large-scale street scenes, existing methods suffer from rapidly escalating per-viewpoint reconstruction costs as scene size increases, leading to significant computational overhead.
After revisiting the conventional pipeline, we identify three key factors accounting for this issue: unnecessary local-to-global transformations, excessive 3D-to-2D projections, and inefficient rendering of distant content. To address these challenges, we propose \textbf{S3R-GS}, a 3DGS framework that \underline{S}treamlines the pipeline for large-scale \underline{S}treet \underline{S}cene \underline{R}econstruction, effectively mitigating these limitations. 
Moreover, most existing street 3DGS methods rely on ground-truth 3D bounding boxes to separate dynamic and static components, but 3D bounding boxes are difficult to obtain, limiting real-world applicability. To address this, we propose an alternative solution with 2D boxes, which are easier to annotate or can be predicted by off-the-shelf vision foundation models. Such designs together make S3R-GS readily adapt to large, in-the-wild scenarios.
Extensive experiments demonstrate that S3R-GS enhances rendering quality and significantly accelerates reconstruction. Remarkably, when applied to videos from the challenging Argoverse2 dataset, it achieves state-of-the-art PSNR and SSIM, reducing reconstruction time to below 50\%—and even 20\%—of competing methods. The code will be released to facilitate further exploration. 

\end{abstract}   
\section{Introduction}
\label{sec:intro}

\begin{figure*}[!t]
\centering
\includegraphics[width=0.95\linewidth]{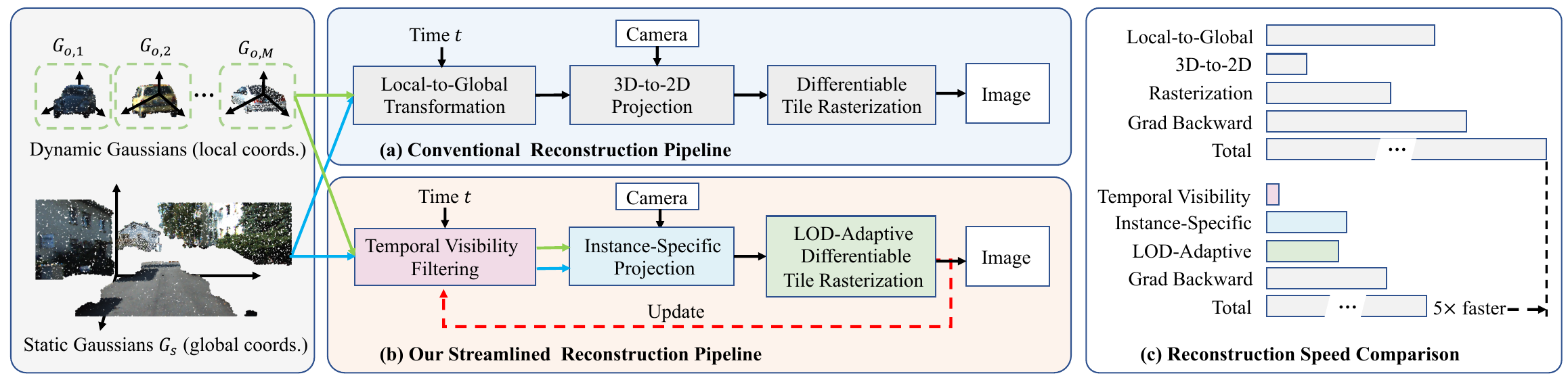}
\vspace{-0.3cm}
\caption{\textbf{Comparison of Reconstruction Pipelines.} \textbf{(a)} In the conventional pipeline, to render the view at timestep $t$, each object's Gaussians are sequentially transformed from their respective local coordinate system to the global coordinate system. Next, all Gaussians in the global coordinate system are projected onto the camera plane. Finally, the Gaussians within the view frustum are rendered using $\alpha$-blending, regardless of distance.  \textbf{(b)} Our streamlined reconstruction pipeline first employs temporal visibility to identify the visible Gaussians at timestep $t$. To avoid unnecessary transformations, we use instance-specific projection matrices to directly project all Gaussians onto the camera plane. Subsequently, we apply the Adaptive LOD method to cull distant Gaussians whose 2D scales are smaller than the LOD threshold. We rasterize the remaining Gaussians with $\alpha$-blending. Finally, we update the temporal visibility using Gaussian visibility obtained by the rendering process. \textbf{(c)} Our pipeline eliminates these redundancies, significantly accelerating the reconstruction process.}
\label{fig:pipeline}
\vspace{-0.2cm}
\end{figure*}

As an emerging topic, street scene reconstructing has attracted a surge of research interest recently due to its significance in applications such as autonomous vehicles, robotics, and VR/AR. It plays a vital role in developing high-quality simulators~\cite{shi2024language,dosovitskiy2017carla,xie2024citydreamer,lee2024semcity}, and also shows great potential in synthesizing driving corner cases~\cite{zhou2024drivinggaussian}. However, unlike the reconstruction of a single object or an indoor environment, it involves a vast number of Gaussian primitives to accurately model the street scene, significantly increasing computational overhead. As the range of driving scenes expands, this challenge becomes even more pronounced, making it a valid problem to be explored.

In the literature, one popular direction towards street scene reconstruction is to develop the pipeline based on Neural Radiance Fields (NeRF) \cite{mildenhall2021nerf}. This kind of method \cite{chenguc,guo2023streetsurf,irshad2023neo,liu2023neural, wang2023neural,zhenxing2022switch,xu2023grid,ost2021neural,lin2023high} utilizes neural networks to represent the scene as a continuous 3D volume and performs volume rendering to generate novel views, remarkably improving the rendering quality of classical reconstruction approaches \cite{furukawa2009accurate, gallup20103d,cornelis20083d}. However, due to the constraints of neural network capacity, NeRF-based methods need large computational overhead when applied to large outdoor scenarios and suffer from artifacts and blurriness in modeling dynamic objects.

Recently, 3D Gaussian Splatting (3DGS) \cite{kerbl20233d} offers an efficient alternative, utilizing explicit 3D Gaussian primitives to represent 3D scenes. This explicit representation facilitates the decomposition of static and dynamic elements, which is a desirable property for scene reconstruction. Additionally, 3DGS accelerates rendering by replacing volume rendering with rasterization-based rendering. Such advantages have led to the trend of utilizing 3DGS in recent works \cite{zhou2024drivinggaussian,yan2024street,zhou2024hugs,huang2024sgaussian, 4dgf} for the reconstruction of street scenes. However, when applied to large-scale street scenes, these methods suffer from rapidly increasing per-viewpoint reconstruction costs due to significant computational redundancy in their pipelines. 
As illustrated in \cref{fig:pipeline} (a), the conventional pipeline generally comprises three steps: transforming each object's Gaussians from local to global frame, projecting all of the Gaussian primitives to the image plane, and rasterizing visible projected Gaussians using $\alpha$-blending method. First, local-to-global transformation is unnecessary for dynamic street scene reconstruction, yet it introduces substantial computational redundancy, particularly for large-scale scenes with hundreds of objects. Second, 3D-to-2D projection is redundant for Gaussians outside the limited field of view. Third, these pipelines apply $\alpha$-blending to all visible projected Gaussians equally, which is inefficient for rendering distant content and increases training costs. These underdesigned steps in the pipeline remarkably impair the reconstruction speed.

To this end, we propose S3R-GS, streamlining the pipeline for large-scale street scene reconstruction. The main improvements of S3R-GS are illustrated in \cref{fig:pipeline} (b).  Specifically, redundant Gaussians are first filtered out based on their temporal visibility. Next, we project dynamic Gaussians of different objects and static Gaussians onto the 2D image plane using the instance-specific projection matrices. We then apply an adaptive level-of-detail (LOD) strategy to cull distant Gaussians, further optimizing rendering efficiency. Finally, the remaining Gaussians are rasterized using $\alpha$-blending, and the temporal visibility of all Gaussians is updated accordingly. Compared to prior pipelines, our pipeline eliminates the need for local-to-global transformations by leveraging the instance-specific projection. Incorporating temporal visibility also mitigates redundant 3D-to-2D projections. Furthermore, the adaptive LOD strategy enables efficient rendering of distant contents. 
As compared in \cref{fig:pipeline} (c), such strategies effectively speed up the reconstruction process.

Besides reconstruction steps, two other factors in the conventional pipeline also affect the applicability of previous methods: (1) the requirements of 3D ground-truth boxes to decompose the dynamic and static objects~\cite{zhou2024drivinggaussian,yan2024street,zhou2024hugs,4dgf}, which are difficult to obtain. (2) the LiDAR-based 3DGS initialization, which cannot adequately cover the tall structures in the street scenes~\cite{caesar2020nuscenes,Geiger2013IJRR}.
To alleviate the cost of 3D annotation, we propose an alternative solution with accessible 2D boxes. However, directly employing 2D boxes along with discrete, learnable pose matrices presents challenges in accurately capturing object motion tracks and estimating object poses at out-of-distribution time steps. Therefore, we introduce a Neural Ordinary Differential Equation (NeuralODE) \cite{neuralode} model, designed to learn continuous motion tracks, thereby facilitating precise scene decomposition.
Moreover, we propose a Bird's Eye View (BEV) semantic initialization augmentation method to supplement the initial 3D Gaussians of tall structures in street scenes.  

We conduct extensive experiments on the KITTI \cite{Geiger2013IJRR}, Argoverse 2 \cite{av2}, and nuScenes \cite{caesar2020nuscenes} datasets to validate the merits of the improved pipeline by S3R-GS. It achieves state-of-the-art performance while reducing reconstruction time to less than 50\%—and in some cases even 20\%—of the methods with conventional pipelines. 

Our contributions can be concluded as follows:
\begin{itemize}
    \item An efficient reconstruction pipeline tailored for large-scale street scenes to reduce the per-viewpoint reconstruction cost as scenes scales up.
    \item An accessible 2D decomposition method enhances adaptability for in-the-wild scenarios, while a BEV-semantic initialization augmentation method improves the reconstruction quality.
    \item A powerful street scene reconstruction framework, which demonstrates superior performance on several publicly available datasets.
\end{itemize}

\section{Related Work}
\label{sec:relatedwork}

\begin{figure*}[!t]
\centering
\includegraphics[width=0.95\linewidth]{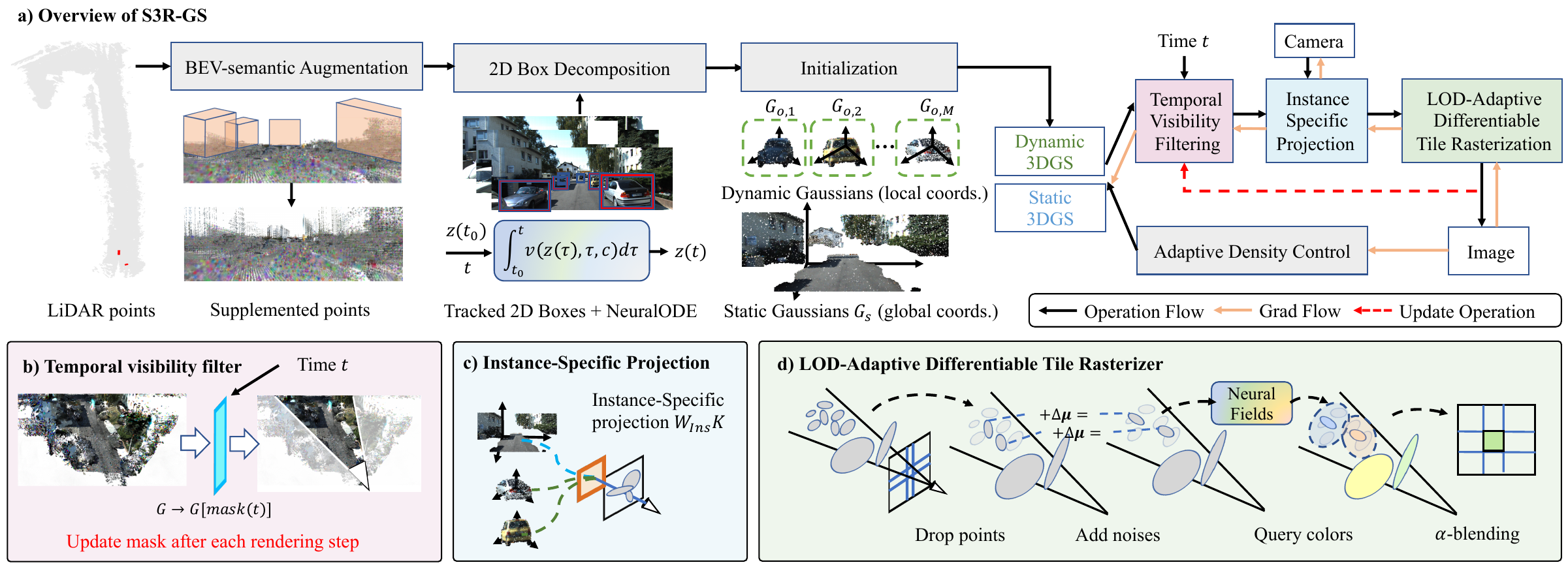}
\vspace{-0.1cm}
\caption{\textbf{S3R-GS Framework.} At the scene modeling stage, S3R-GS first leverages a BEV-semantic initialization augmentation method to supplement the points of tall structures in street scenes. Next, tracked 2D boxes to distinguish between dynamic and static elements, integrating a NeuralODE model for precise, continuous object poses. These designs enable S3R-GS to generalize effectively to in-the-wild scenarios. 
During scene reconstruction, 
S3R-GS identifies visible Gaussians at each time step and projects them onto the 2D image plane using instance-specific projection matrices. 
To efficiently render distant content, S3R-GS employs an adaptive strategy that (1) filters out distant 3D Gaussians with small projected 2D scales, (2) randomly culls Gaussians based on depth, (3) introduces noisy offsets to the remaining distant Gaussians, and (4) queries their colors from a distance-aware neural field. After that, the remaining Gaussians are rasterized using $\alpha$-blending. Our pipeline reduces computational redundancy, significantly lowering per-viewpoint reconstruction costs.}
\label{fig:method}
\vspace{-0.2cm}
\end{figure*}

\subsection{NeRF for Street Scene Reconstruction}

By leveraging multi-layer perceptrons (MLPs) and differentiable volume rendering, Neural Radiance Fields (NeRF)~\cite{mildenhall2021nerf,barron2021mip,barron2022mip,muller2022instant,barron2023zip,fridovich2022plenoxels,garbin2021fastnerf} has demonstrated remarkable performance in 3D reconstruction and novel view synthesis.
Many studies \cite{chenguc,guo2023streetsurf,irshad2023neo,liu2023neural,lu2023urban,ost2022neural,rematas2022urban,tancik2022block,wang2022neural,martin2021nerf,turki2022mega,wang2023neural,zhenxing2022switch,xu2023grid,ost2021neural,lin2023high,lin2022efficient,park2021hypernerf,song2023nerfplayer} have adapted NeRF for urban scenes, though most focus on static environments. For dynamic scenes, NSG \cite{ost2021neural} decomposes scenes into graphs, EmerNeRF \cite{yang2023emernerf} integrates spatial-temporal representations, and DNMP \cite{lu2023urban} employs deformable neural meshes with LiDAR priors.
Despite efforts to accelerate NeRF training \cite{garbin2021fastnerf,barron2022mip}, large-scale dynamic urban reconstruction remains challenging due to high computational costs and real-time rendering constraints. Additionally, NeRF struggles with explicit scene modeling and fine-detail reconstruction, limiting its effectiveness for large-scale street scenes.

\subsection{3DGS for Street Scene Reconstruction}

Compared to NeRF, 3D Gaussian Splatting (3DGS) \cite{aliev2020neural,dai2020neural,kopanas2021point,kerbl20233d} offers faster training, more efficient rendering, and higher reconstruction quality by representing scenes with explicit anisotropic 3D splats, reducing unnecessary computations in empty regions.
While 3DGS is initially designed for static scene reconstruction, recent studies  \cite{luiten2024dynamic, yang2024deformable, wu20244d} have extended 3DGS to room-level dynamic scene reconstruction. 
At the same time, some studies also extended 3DGS to dynamic urban scene modeling. DrivingGaussian \cite{zhou2024drivinggaussian} reconstructs large-scale urban scenes by separating static and dynamic elements using tracked 3D bounding boxes and employing incremental static 3D Gaussian to model the large-scale scenes in the depth range, but this approach complicates the reconstruction process and significantly increases the reconstruction time. StreetGaussian \cite{yan2024street} models dynamic urban scenes with optimizable tracked poses and a 4D spherical harmonics model for the dynamic appearance. HUGS \cite{zhou2024hugs} employs 3D Gaussian Splatting for comprehensive urban scene understanding, which jointly optimizes geometry, appearance, semantics, and motion. To reduce memory costs, 4DGF \cite{4dgf} utilizes 3D Gaussians as a geometric scaffold and introduces neural fields for complex appearance modeling. 

Despite impressive progress, these methods still struggle with scalability, as training time per viewpoint increases with scene size, limiting their efficiency in large-scale street scene reconstruction. Therefore, in this work, we attempt to streamline the pipeline and alleviate this problem.
\section{Method}

In this work, we present S3R-GS, a 3DGS framework that streamlines the pipeline for large-scale street scene reconstruction. This section starts with the preliminary of the conventional street scene reconstruction pipeline (\cref{sec:preliminary}), which typically consists of two stages: scene modeling and scene reconstruction. Then, we present our improvements in the scene modeling stage (\cref{sec:scene modeling}). After that, we detail our streamlined reconstruction pipeline (\cref{sec:scene reconstruction}), which enhances efficiency and scalability in the scene reconstruction stage. An overview of S3R-GS is shown in \cref{fig:method}.

\subsection{Preliminary}
\label{sec:preliminary}
We build our S3R-GS upon the commonly used reconstruction pipeline that is widely adopted in recent works \cite{zhou2024drivinggaussian,yan2024street,zhou2024hugs,4dgf}. Specifically, in the scene modeling stage, given images $I \in {{\mathbb{R}}^{T \times H \times W \times 3}}$, camera extrinsic $W \in SE(3)$ and intrinsics $K \in \mathbb{R}^{T \times 3 \times 3}$, LiDAR points $P \in {{\mathbb{R}}^{M \times 3}}$ and 3D bounding box trajectories $B \in {{\mathbb{R}}^{K\times T \times 6}}$, 3D Gaussians $G$ are first initialized by LiDAR points, and then decomposed by 3D bounding boxes into the static Gaussians $G_s$ for the static scene and dynamic Gaussians $G_{o,i}$ for each object $O_i \in \{O_1,O_2,...,O_K\}$. In the scene reconstruction stage, since the positions of objects vary over time in dynamic street scenes, conventional reconstruction pipelines often transform dynamic Gaussians $G_{o,i}$ of each object $O_i$ from their respective local coordinate systems to the global coordinate system to ensure that different objects are aligned within a global space.
Second, all 3D Gaussians $G$ in global space are projected to the 2D image plane with camera extrinsic $W_t$ and intrinsic $K_t$, and 2D positions $\mu'$ and 2D covariances $\Sigma'$ of visible Gaussians are computed: 
\begin{equation}
\begin{aligned}
    \mu' &=K_tW_t\mu \\
    \Sigma' &=JW_t\Sigma W_t^T J^T,
\end{aligned}
\end{equation}
where $J$ is the Jacobian matrix of $K_t$. $\mu$ and $\Sigma$ denote the positions and covariance matrices of the 3D Gaussians, respectively. Simultaneously, the color of 2D Gaussians is calculated based on a set of spherical harmonic (SH) coefficients \cite{zhou2024drivinggaussian,yan2024street,huang2024sgaussian}. A variant \cite{4dgf} uses 3D Gaussians only as a geometry scaffold and queries the color of 3D Gaussians from neural fields, conditioned on the positions of the 3D Gaussians and the camera direction. Finally, the color $C$ of each 2D pixel is rasterized by point-based $\alpha$-blending:
\begin{equation}
{
    C=\sum_{i\in N}c_i \alpha_i \prod_{j=1}^{i-1}(1-\alpha_j),
\label{eq:color}}
\end{equation}
where $c_i$ and $\alpha_i$ denote the color and opacity of the 2D Gaussians covering the pixel, respectively.  

\subsection{Improvements in Scene Modeling Stage}
\label{sec:scene modeling}
We make improvements in the scene modeling stage through a BEV-semantic initialization augmentation and an efficient 2D boxes decomposed method.

\vspace{+0.1cm}
\noindent \textbf{BEV-Semantic initialization augmentation.}
Due to the limited vertical view field~\cite{caesar2020nuscenes,Geiger2013IJRR}, it is difficult for LiDAR sensors to capture high-rise buildings and other elevated structures, affecting 3DGS initialization.  To alleviate it, we propose a BEV-semantic augmentation method as supplementary. Specifically, we first obtain the 2D semantic image $S\in \mathbb{Z}^{H \times W \times 1}$ of each image by \cite{liu2023grounding}. Then, we project point clouds $P$ onto the 2D semantic image to obtain their pixel coordinates $UV \in \mathbb{Z}^{M\times 2}$ and semantic labels $S \in \mathbb{Z}^{M \times 1}$. Then, we select the points $P' \in \mathbb{R}^{M'\times 3}$ with specific semantic labels, such as ``tree" or ``building", and obtain their BEV positions $XY \in \mathbb{R}^{M'\times2}$. To enhance scene initialization, we partition the scene into uniform grids based on XY coordinates and identify grids containing points $P'$. For each occupied grid $(x,y)$, we re-initialize the points $P_{aug}=\{(x,y,\Delta z),(x,y,2\Delta z),...,(x,y,h)\}$ at a interval $\Delta z$ along the z-axis, where $h$ is the predefined height. Finally, we project re-initialized points $P_{aug}$ back into images and retain only the points $P_{aug}'$ within the image frustum. The augmented points $P'_{aug}$ are merged with the original points $P$ to complete the scene initialization.

\noindent \textbf{2D boxes for scene decomposition.}

Most current street scene reconstruction methods\cite{yang2023emernerf,huang2024sgaussian} depend on tracked 3D bounding boxes to distinguish static from dynamic elements, limiting their applicability in uncontrolled, real-world scenarios.
To this end, we proposed an alternative solution with 2D boxes, which is easier to obtain. 

Particularly, we introduce a NeuralODE \cite{neuralode} model that treats an object's motion trajectory as a continuous ordinary differential equation to learn its accurate 3D pose. 
Specifically, given images $I \in \mathbb{R}^{T\times H \times W}$, LiDAR points $P\in R^{M\times 3}$ and 2D Boxes $B \in \mathbb{R}^{K \times T \times 4}$, we first obtain the 2D masks of objects from 2D boxes using the segment model \cite{liu2023grounding,kirillov2023segany} and project the point clouds captured at each time $t$ onto the corresponding image $I_t$ to extract the points within 2D masks. We then compute the mean 3D position of the points within each 2D mask, treating it as the estimated 3D location of the object at time $t$, denoted as $XYZ_t\in \mathbb{R}^3$. Through this process, we obtain a coarse trajectory $TXYZ \in \mathbb{R}^{T \times 3}$ for each object and initialize its rotation angle $R$ to zero. During reconstruction, we input both the time $t$ and the object's coarse 3D position at time $t$ into NeuralODE to obtain the refined 3D position and rotation angle at that time step. To enable the network to distinguish between different object instances with different categories and different speeds, we introduce a instance embedding $C\in \mathbb{R}^{|K| \times D}$. Specifically, we assign a unique embedding $c \in \mathbb{R}^D$ to each object and incorporate it as an additional input to the NeuralODE. This allows the model to account for instance-specific motion patterns, improving the accuracy of trajectory estimation. NeuralODE models an object's state evolution over time using an ordinary differential equation parameterized by a neural network. Specifically, we define the object's state at time $t$ as $\mathbf{z}(t) = [ \Delta XYZ_t + XYZ_t,\ R_{t}] \in \mathbb{R}^{6}.$
$\Delta XYZ_t$ denotes the offset of the object as time $t$, learned by NeuralODE. $R_t$ denotes its rotation parameters. The state evolution is modeled as:
\begin{equation}
    \frac{d\mathbf{z}(t)}{dt} = f(\mathbf{z}(t), t, c),
\end{equation}
where $f(\cdot)$ is the NeuralODE network, which learns the object's motion dynamics. $c$ is the instance embedding of the object. Given an initial state $\mathbf{z}(t_0)$, the state at any future time $t$ can be computed via numerical integration:
\begin{equation}
    \mathbf{z}(t) = \mathbf{z}(t_0) + \int_{t_0}^{t} f(\mathbf{z}(\tau), \tau, c) d\tau.
\end{equation}
This continuous formulation allows NeuralODE to learn smooth and physically plausible object trajectories, capturing motion continuity that traditional discrete-time models often fail to represent. 

\subsection{Streamlined Reconstruction Stage}
\label{sec:scene reconstruction}
Existing street scene reconstruction pipelines \cite{zhou2024drivinggaussian,huang2024sgaussian,zhou2024hugs,yan2024street,4dgf} often suffer from significant computational redundancy when applied to large-scale reconstruction, resulting in a substantial decrease in reconstruction speed. Three main computational redundancies in their pipelines contribute to this issue: unnecessary local-to-global coordinate transformations,
redundant 3D-to-2D projections,
and inefficient rendering of distant content.

To reduce the aforementioned computational redundancies, we propose a novel street reconstruction pipeline tailored for large-scale street scenes, building upon the conventional pipeline \cite{zhou2024drivinggaussian,zhou2024hugs,yan2024street,4dgf}. In our method, each Gaussian is assigned a 3D position $\mu \in \mathbb{R}^3$, a 3D covariance $\Sigma \in \mathbb{R}^3$, an opacity $\alpha \in \mathbb{R}$, a temporal visibility $v \in \mathbb{R}^{2}$ and a point life $l \in \mathbb{R}^{2}$. Additionally, dynamic Gaussian is associated with an instance $ID \in \mathbb{Z}$, indicating their corresponding object. Temporal visibility records the time interval during which a Gaussian remains visible, including its appearance time and disappearance time. To reduce the memory footprint, we follow 4DGF \cite{4dgf} to adapt the neural fields $NeurF(*)$ to learn Gaussian colors. At each time $t$ rendering, our pipeline proceeds as follows: First, we select the visible Gaussians $G_t$ at time $t$ based on their temporal visibility. Second, the selected Gaussians $G_t$ are projected onto the 2D image plane using instance-specific camera parameters $TW^{K \times 3 \times 3}$, eliminating the need for local-to-global transformations. After projecting Gaussians to the image plane, we select the visible Gaussians $G'_t$ in the view frustum and obtain the new visible mask $M_t \in \mathbb{B}^{N}$. Third, after obtaining the 2D scales $\Sigma'$ of visible Gaussians $G'_t$ from the projection step, we cull Gaussians $G'_{t,\Sigma' \leq r}$ with 2D scales below a predefined level-of-detail (LOD) threshold $r$ to a subset $G''_{t,\Sigma' \leq r}$ with the adaptive LOD strategy. After culling, we query the colors of the final visible 3D Gaussians $\{G'_{t,\Sigma' > r},G''_{t,\Sigma' \leq r}\}$ within the current view frustum and proceed with rasterization as \cref{eq:color}. The temporal visibility of all Gaussians are also updated by $M_t$.

\vspace{+0.1cm}
\noindent \textbf{Instance-specific projection.} 
At each rendering time $t$ of the conventional pipeline, the Gaussians $G_{o,i}$ of each object $O_i$ will be transformed from their local coordinate systems to the global coordinate systems using transformation $W_{t,i2g} \in \mathbb{R}^{4\times 4}$. To avoid unnecessary computations, for each time $t$ with a camera $c=\{W_t,K_t\}$, we predefine an instance-specific camera for each object. The intrinsic parameters of these cameras remain unchanged, while the extrinsic parameters are set as $W_{t,i}=W_tW_{t,i2g}$. During rendering, we simply select the corresponding cameras based on the Gaussian's instance ID, thereby eliminating the need for explicit coordinate transformations. This modification is simple yet significantly improves the reconstruction speed.

\begin{figure*}[!t]
\centering
\includegraphics[width=0.9\linewidth]{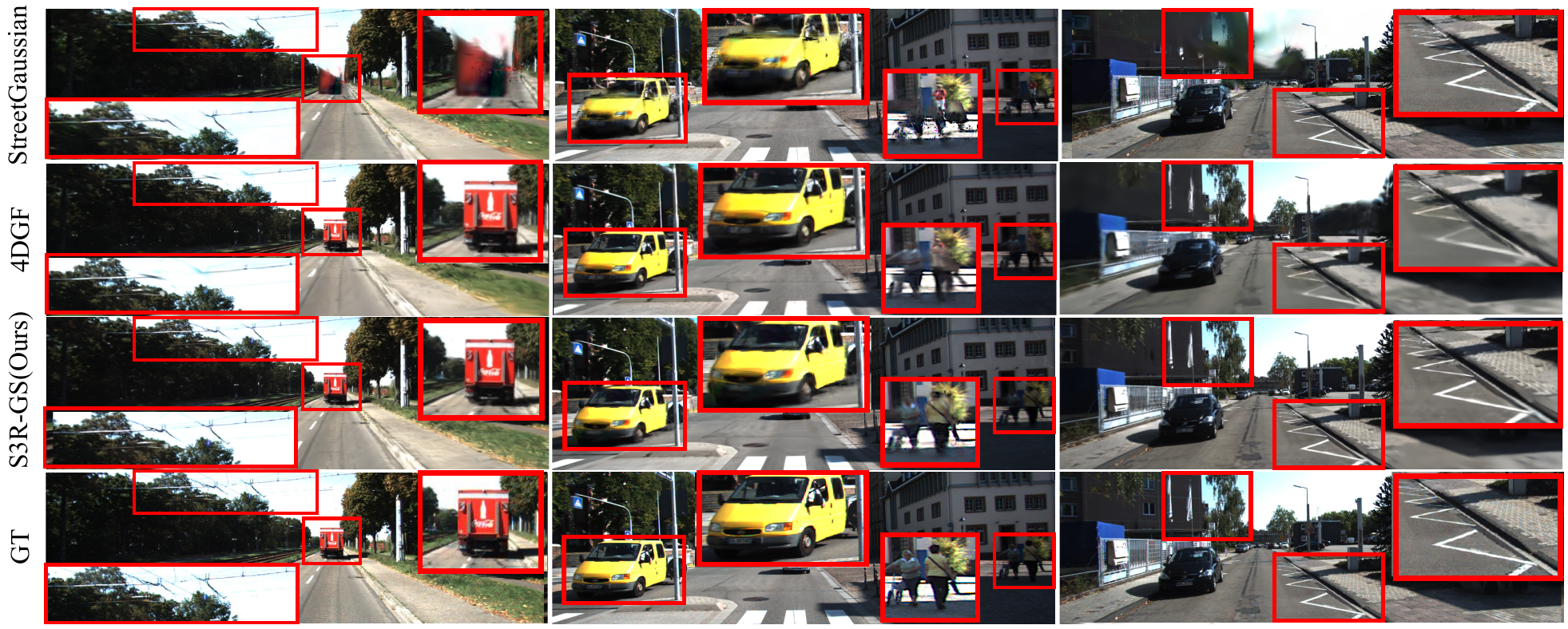}
\vspace{-0.3cm}
\caption{\textbf{Qualitative Comparisons on the KITTI Dataset.} StreetGaussian \cite{yan2024street} fails to accurately model the details of vehicles and pedestrians. 4DGF \cite{4dgf} struggle to model the details of large-scale scenes. In contrast, our method not only reconstructs large-scale scenes with full detail but also achieves high-quality reconstructions of vehicles and pedestrians.}
\label{fig:vis}
\vspace{-0.5em}
\end{figure*}

\vspace{+0.1cm}
\noindent \textbf{Temporal separation.} 
In the conventional pipeline, after the local-to-global transformation step, all 3D Gaussians $G$ in the global coordinate system will be projected onto the 2D image plane of the current viewpoint, and then the Gaussians out of the view frustum will be filtered out. To reduce the redundant 3D-to-2D projections, we introduce temporal visibility to filter out the invisible Gaussians at the current time step before the projection. Specifically, we first normalize the rendering time of the whole scene to [-1,1]. To avoid unnecessary computations, we assign each Gaussian a temporal visibility $v \in \mathbb{R}^2$, defined by an appearance time $v_s \in \mathbb{R}$ and a disappearance time $v_e \in \mathbb{R}$, which are initially set to -1 and 1, respectively. We also assign a point life attribute $l \in \mathbb{R}^{2}$ to each Gaussian, defined by an appearance time $l_s \in \mathbb{R}$ and a disappearance time $l_e \in \mathbb{R}$, which are initially set to 1 and -1, respectively. During rendering at time $t$, instead of projecting all Gaussians onto the image plane to select a visible subset, we directly select the visible Gaussians $G'$ whose visibility intervals encompass $t$, denoted as $t_s\leq t\leq t_e$. Then, we only need to project visible Gaussians $G'$ onto the 2D image plane, thereby reducing computational redundancy. After each rendering, we can obtain the actual visible Gaussians mask $M_t \in \mathbb{B}^{N}$ and update the point life of each Gaussians $i$:
\begin{equation}
\begin{aligned}
    l_{s,i} &= min(l_{s,i},t)\ if M_t[i]==True.\\
    l_{e,i} &= max(l_{e,i},t)\ if M_t[i]==True.
\end{aligned}
\end{equation}
After iterating through all training times, we update the temporal visibility $v$ of each Gaussian by its point life $l$:
\begin{equation}
        t_{s} = l_{s}-0.1,\ t_{e} = l_{e}+0.1.
\end{equation}
Newly added points are initially set to be visible at all time steps. In addition, we periodically reset the temporal visibility $t$ of all Gaussians to be visible to maintain consistency across multiple viewpoints.

\vspace{+0.1cm}
\noindent \textbf{Adaptive LOD.}
In the conventional pipeline, after the 3D-to-2D projection step, the color of all visible Gaussians $G'$ will be calculated, and all of them will be rasterized regardless of their distances and scales, thereby resulting in inefficient rendering of distant content for large-scale street scenes. To address this, we propose an adaptive level-of-detail (LOD) strategy that adaptively selects which Gaussians to render, reducing redundant computations for distant content. Specifically, after obtaining the 2D positions $\mu'$, 2D scales $\Sigma'$, and depths $d$ of visible Gaussians $G'_t$ from the projection step, we filter out Gaussians $G'_{t,\Sigma' \leq r}$ with scales below a predefined level-of-detail (LOD) threshold $r$, such as 4 pixels. For these Gaussians $G'_{t,\Sigma' \leq r}$, we probabilistically cull a subset based on their depths, with farther Gaussains having a higher cull probability. To obtain the average color of surrounding regions, we apply a small noisy offset $\Delta \mu$ to the remaining Gaussians $G''_{t,\Sigma' \leq r}$ of $G'_{t,\Sigma' \leq r}$. After culling, we query the colors of the final visible 3D Gaussians $\{G'_{t,\Sigma' > r},G''_{t,\Sigma' \leq r}\}$ within the current view frustum and proceed with rasterization as \cref{eq:color}. This process is denoted as:
\begin{equation}
    \begin{aligned}
    &p = p_{max} + (p_{max} - 10^{-2}) \cdot \min(0, \frac{d - D}{D}), \\
    &M_{\text{LOD}} = \text{Bernoulli}(p) \leq 0, \\
    &\mu \leftarrow \mu[M_{\text{LOD}}],\ d \leftarrow d[M_{\text{LOD}}],\\
    &\mu \leftarrow \mu + [\Delta x, \Delta y, \Delta z]\cdot normalize(d) \cdot \mathcal{N}(0,1), \\
    &c = NeurF_{sta}(\mu, d, dir, emb(t)), \\
    &c = NeurF_{dyn}(\mu, d, dir, emb(t), class), \\
\end{aligned}
\end{equation}
where $p$ represents the drop probability of each Gaussian, $p_{max}$ denotes the predefined maximum drop probability. The binary mask $M_{\text{LOD}}$ is sampled from a Bernoulli distribution with probability $p$, determining whether a given Gaussian component is retained. $\Delta x, \Delta y, \Delta z$ are the predefined position offset scale. A noisy offset related to the depth is added to the Gaussian. For static Gaussians, the static neural field $NeurF_{sta}$ computes color $c$ based on position $\mu$, depth $d$, viewing direction $dir$, and time-dependent appearance embeddings. For dynamic Gaussians, $NeurF_{dyn}$ additionally incorporates the class embeddings $class$. Introducing the depth as an input to neural fields enables the network to automatically learn color variations across different LODs.

\section{Experiment}
\subsection{Experimental Setup}

\noindent\textbf{Datasets.} We empirically verify the efficiency of our S3R-GS framework for dynamic large-scale street scene reconstruction in comparison with state-of-the-art approaches on three datasets, i.e., Argoverse 2 \cite{av2}, KITTI \cite{Geiger2013IJRR}, and nuScenes \cite{caesar2020nuscenes}. We first validate S3R-GS on the challenging large-scale street scene NVS benchmark \cite{4dgf} of Avgoverse 2, which consists of multi-sequence scenarios and seven camera views. Second, we evaluate S3R-GS on two large-scale street scenes selected from KITTI. Moreover, we also evaluate our methods on the widely used NVS benchmark \cite{turki2023suds,yan2024street,4dgf} of KITTI, which consists of three small scenes. Additionally, we also evaluate S3R-GS on the NVS benchmark \cite{zhou2024drivinggaussian} of nuScenes dataset, which is a large public autonomous driving dataset, consisting of six cameras. Following recent works \cite{yan2024street,4dgf}, we adapt PSNR, SSIM \cite{wang2004image} and LPIPS \cite{zhang2018unreasonable} as novel view synthesis metrics.

\noindent\textbf{Implementation Details}
We utilize LiDAR points prior to precise geometric initialization and employ a voxel grid downsampling method to reduce LiDAR points, with the grid size set to 0.15 meters. 

To facilitate the growth and pruning of 3D Gaussians, we adopt an adaptive density control mechanism proposed in \cite{4dgf}. For non-rigid objects, we also use a deformable neural field to learn the ego offsets of non-rigid object Gaussian positions at each time $t$. All the experiments are conducted on NVIDIA V100 GPUs with 32 GB of memory. Details of our implementation will be shown in the supplementary.

\subsection{Results and Comparisons}
\begin{table*}[htbp]
    \renewcommand{\arraystretch}{1.05}
  \centering
   \resizebox{0.97\textwidth}{!}{
  \begin{tabular}{l|ccc|ccc|cccc}
\hline
\hline
\rowcolor[gray]{.92} & \multicolumn{3}{c|}{Residential} & \multicolumn{3}{c|}{Downtown} & \multicolumn{4}{c}{Mean}\\ 
 \rowcolor[gray]{.92} Method & PSNR $\uparrow$ & SSIM $\uparrow$ & LPIPS $\downarrow$ & PSNR $\uparrow$ & SSIM $\uparrow$ & LPIPS $\downarrow$ & PSNR $\uparrow$ & SSIM $\uparrow$ & LPIPS $\downarrow$ & Rec. Time $\downarrow$ \\ \hline
SUDS \cite{turki2023suds} & 21.76 & 0.659 & 0.556 & 19.91 & 0.665 & 0.645 & 20.84 & 0.662 & 0.601 & - \\
ML-NSG \cite{fischer2024multi} & 22.29 & 0.678 & 0.523 & 20.01 & 0.681 & 0.586 & 21.15 & 0.680 & 0.555 & 49.10 h  \\  
4DGF \cite{4dgf} & 25.78 & 0.772 & \textbf{0.405} & 24.16 & 0.772 & 0.488 & 24.97 & 0.772 & 0.447 & 54.39 h \\ 
\rowcolor{tabblue!10} S3R-GS (Ours) & \textbf{26.27} & \textbf{0.775} & 0.406 & \textbf{25.08} & \textbf{0.785} & \textbf{0.463} & \textbf{25.68} & \textbf{0.780} & \textbf{0.435} & \textbf{26.71 h}  \\ \hline \hline
\end{tabular}
}
\vspace{-0.25cm}
  \caption{\textbf{Comparison Results on Large-Scale Street Scenes from Argoverse 2 \cite{av2}.} Our method surpasses the state-of-the-art approach in reconstruction quality while requiring less than half the reconstruction time.}
  \label{tab:av2}
  \vspace{-0.25cm}
\end{table*}

\begin{table*}
    \renewcommand{\arraystretch}{1.05}
  \centering
\resizebox{0.97\textwidth}{!}{
\begin{tabular}{l|cccc|cccc}
\hline
\hline
\rowcolor[gray]{.92} & \multicolumn{4}{c|}{KITTI 0009 full lengths [75\%]} & \multicolumn{4}{c}{KITTI 0020 full lengths [75\%]}\\ 
 \rowcolor[gray]{.92} Method & PSNR $\uparrow$ & SSIM $\uparrow$ & LPIPS $\downarrow$ & Rec. Time $\downarrow$  & PSNR $\uparrow$ & SSIM $\uparrow$ & LPIPS $\downarrow$ & Rec. Time $\downarrow$ \\ \hline
StreetGaussians \cite{yan2024street} & 22.17 & 0.703 & 0.297 & 5.80 h & 22.89 & 0.789 & 0.288 & 6.53 h  \\
ML-NSG \cite{fischer2024multi} & 20.50 & 0.589 & 0.466 & 18.17 h  & 21.35 & 0.683 & 0.439 & 19.21 h \\
4DGF \cite{4dgf} & 22.64 & 0.706 & 0.301 & 15.69 h & 23.08 & 0.787 & 0.311 & 18.04 h  \\ 
\rowcolor{tabblue!10}S3R-GS (Ours) & \textbf{23.86} & \textbf{0.746} &  \textbf{0.274} & \textbf{3.02 h}  & \textbf{24.96} & \textbf{0.828} & \textbf{0.260} & \textbf{3.37 h}  \\ \hline \hline
\end{tabular}
}
\vspace{-0.25cm}
  \caption{\textbf{Comparison Results on Large-Scale Street Scenes of KITTI \cite{Geiger2013IJRR}.} Our method not only outperforms the state-of-the-art in reconstruction quality but also achieves a 5$\times$ acceleration in reconstruction speed.}
  \label{tab:kitti_large}
  \vspace{-0.3cm}
\end{table*}

\begin{table*}
\renewcommand{\arraystretch}{1.05}
  \centering
  \resizebox{0.95\textwidth}{!}{
  \begin{tabular}{l|ccc|ccc|ccc}
\hline
\hline
\rowcolor[gray]{.92} & \multicolumn{3}{c|}{KITTI [25\%]} & \multicolumn{3}{c|}{KITTI [50\%]} & \multicolumn{3}{c}{KITTI [75\%]}\\ 
 \rowcolor[gray]{.92} Method & PSNR $\uparrow$ & SSIM $\uparrow$ & LPIPS $\downarrow$ & PSNR $\uparrow$ & SSIM $\uparrow$ & LPIPS $\downarrow$ & PSNR $\uparrow$ & SSIM $\uparrow$ & LPIPS $\downarrow$ \\ \hline
 NSG \cite{ost2021neural} & 20.00 & 0.632 & 0.281 & 21.26 & 0.659 & 0.266 & 21.53 & 0.673 & 0.254 \\
SUDS \cite{turki2023suds} & 20.76 & 0.747 & 0.198 & 23.12 & 0.821 & 0.135 & 22.77 & 0.797 & 0.171 \\
MARS \cite{wu2023mars} & 23.23 & 0.756 & 0.177 & 24.00 & 0.801 & 0.164 & 24.23 & 0.845 & 0.160 \\  
StreetGaussians \cite{yan2024street} & 24.53 & 0.824 & 0.090 & 25.52 & 0.841 & 0.084 & 25.79 & 0.844 & 0.081 \\
ML-NSG \cite{fischer2024multi} & 26.51 & 0.887 & 0.060 & 27.51 & 0.898 & 0.055 & 28.38 & 0.907 & 0.052 \\
4DGF \cite{4dgf} & 29.08 & 0.908 & 0.036 & 30.55 & 0.931 & 0.028 & 31.34 & 0.945 & 0.026 \\ 
\rowcolor{tabblue!10} S3R-GS (Ours) & \textbf{29.76} & \textbf{0.933} & \textbf{0.030} & \textbf{31.01} & \textbf{0.943} & \textbf{0.025} & \textbf{31.49} & \textbf{0.946} & \textbf{0.024} \\ 
\hline \hline
\end{tabular}
}
\vspace{-0.2cm}
  \caption{\textbf{Comparison Results on the Novel Synthesis Benchmark of KITTI \cite{Geiger2013IJRR}.} Our method achieves state-of-the-art performance under different training view fractions.}
  \label{tab:kitti_small}
  \vspace{-0.3cm}
\end{table*}

\begin{table}
\renewcommand{\arraystretch}{1.05}
  \centering
  \small
  \resizebox{0.95\linewidth}{!}{
  \begin{tabular}{l|ccc}
    \hline \hline
    \rowcolor[gray]{.92}  & \multicolumn{3}{c}{Novel view synthesis} \\
     \rowcolor[gray]{.92} Method & PSNR $\uparrow$ & SSIM $\uparrow$ & LPIPS $\downarrow$ \\
        \hline
        Mip-NeRF360 \cite{barron2022mip} & 22.61 &  0.688 &  0.395 \\
        Urban-NeRF \cite{chenguc} & 20.75& 0.627& 0.480\\
        SUDS \cite{turki2023suds} & 21.26 &  0.603 & 0.466\\
        EmerNeRF \cite{yang2023emernerf} & 24.23 & 0.845 & 0.160 \\
        DrivingGaussian \cite{zhou2024drivinggaussian} & 28.74 & 0.865 & 0.237 \\
        \rowcolor{tabblue!10} S3R-GS (Ours) & \textbf{29.21} & \textbf{0.868} & \textbf{0.139} \\
    \hline \hline
  \end{tabular}
  }
  \vspace{-0.2cm}
  \caption{\textbf{Comparison Results on nuScenes \cite{caesar2020nuscenes} Datasets.} Our method surpasses all competing approaches.}
  \vspace{-0.3cm}
  \label{tab:nuscene}
\end{table}
\paragraph{Comparisons of Novel Views Synthesis on Avgoverse 2.} 
\cref{tab:av2} summarizes the performance comparisons of different methods for novel views synthesis dataset. Overall, the results across most metrics consistently indicate that our S3R-GS achieves superior performances against the state-of-the-art models. Notably, our S3R-GS reduces reconstruction time by 27.68 hours, achieving up to a 50.9\% speedup compared to the state-of-the-art work 4DGF. The results of Avgoverse 2 highlight the effectiveness of our S3R-GS in large-scale dynamic street scene reconstruction. 

\paragraph{Comparisons of Novel Views Synthesis on KITTI.} 
\cref{tab:kitti_large} shows the performance comparison on large-scale scenes of KITTI datasets for novel views synthesis. Compared to the baselines, our S3R-GS achieves better reconstruction quality in all metrics across all evaluated methods, while significantly reducing reconstruction time. Specifically, our S3R-GS reduces the reconstruction time of the KITTI 0009 scene by 12.67 hours, achieving up to 5$\times$ faster reconstruction speed. For the KITTI 0020 scene with 127 vehicles, our S3R-GS reduces the reconstruction time by 14.67 hours, achieving up to 5.3$\times$ faster reconstruction speed. The reconstruction results further corroborates the effectiveness of our S3R-GS in large-scale dynamic street scene reconstruction. Furthermore, to evaluate the generalizability of our approach, we also assess S3R-GS on a widely used novel view synthesis benchmark, which consists of relatively small-scale scenes, each averaging only 63 frames. As shown in \cref{tab:kitti_small}, our method still achieves state-of-the-art performance, outperforming all baselines, further validating the effectiveness of our approach. 

Moreover, we also show the qualitative results of our method and StreetGaussian \cite{yan2024street}, 4DGF \cite{4dgf} in \cref{fig:vis}. StreetGaussian \cite{yan2024street} fails to accurately model the details of vehicles and pedestrians, leading to blurry and distorted results. 4DGF \cite{4dgf} produces blurry renderings of scene details when applied to large-scale street scenes. In contrast, our method achieves high-quality reconstructions of vehicles and pedestrians, preserving fine details even in large-scale street scenes.

\paragraph{Comparisons of Novel Views Synthesis on nuScenes.} 
Different from the KITTI dataset, nuScenes dataset has more spare views and more complex scenes, making the reconstruction more challenging. We compare our method with several works including NeRF-based methods Mip-NeRF360 \cite{barron2022mip}, Urban-NeRF \cite{chenguc}, SUDS \cite{turki2023suds}, EmerNeRF \cite{yang2023emernerf} and 3DGS-based methods DrivingGaussian \cite{zhou2024drivinggaussian}, StreetGaussian \cite{yan2024street}. As shown in \cref{tab:nuscene}, our method achieves the best performance on all metrics in the novel view synthesis task on nuScenes datasets. 

\subsection{Ablation Study}

\paragraph{Non-Scalability of Per-Viewpoint Reconstruction Cost.}
\begin{figure}[!t]
\centering
\includegraphics[width=0.8\linewidth]{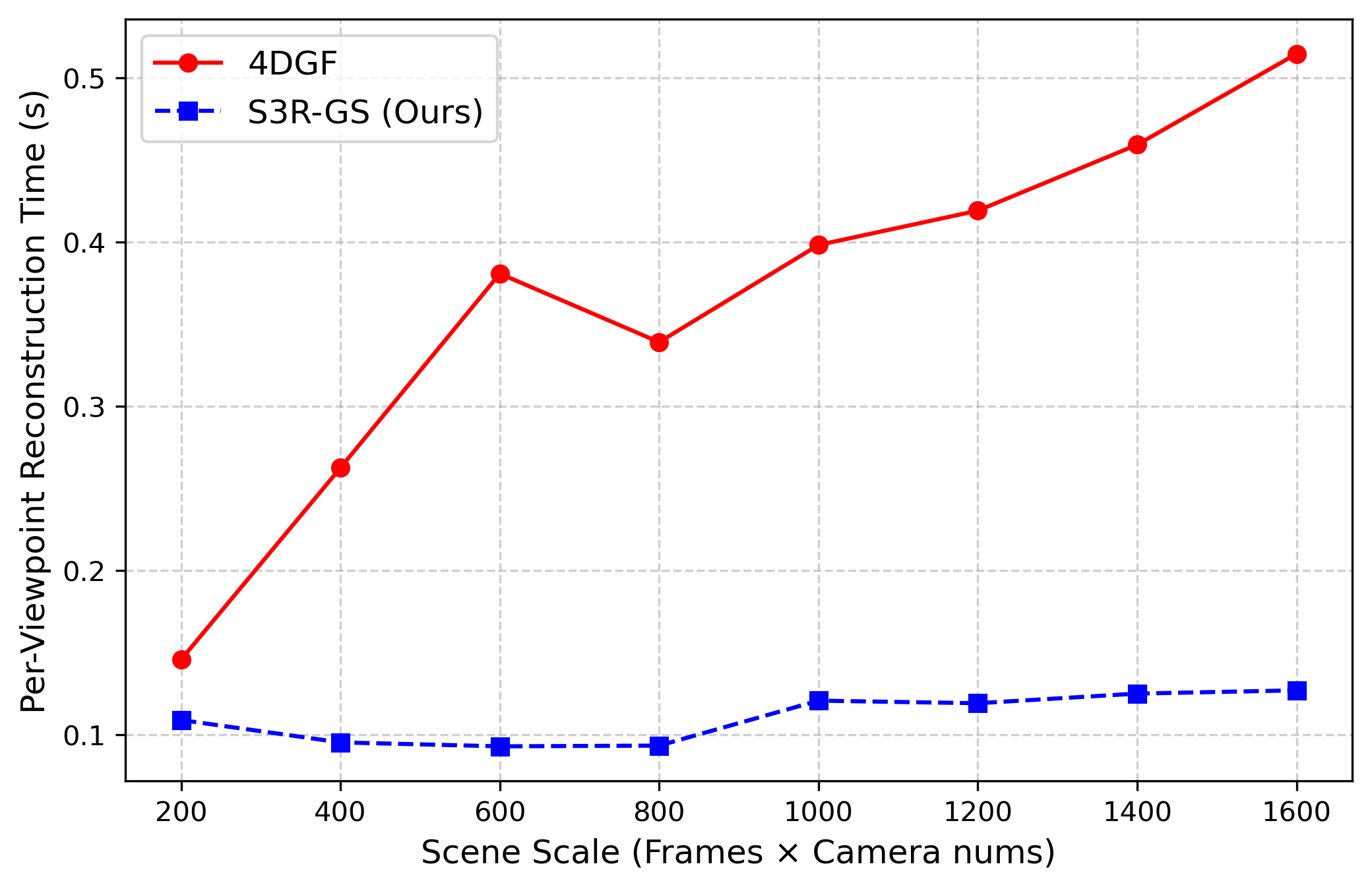}
\caption{\textbf{Scalability Comparison of Per-viewpoint Reconstruction Times}. As the scene scales up, the per-viewpoint reconstruction cost of 4DGF rises rapidly, whereas our method remains nearly constant.}
\label{fig:scale_up}
\vspace{-1.2em}
\end{figure}
To validate that the per-viewpoint rendering cost of our pipeline does not increase rapidly with scene scale, we compare pre-viewpoint reconstruction times across scenes of different sizes. We divide the KITTI 09 scene into sub-scenes of varying frame scales based on time. For instance, the 200-frame sub-scene corresponds to the first 1/8 duration of KITTI 09. \cref{fig:scale_up} presents a comparison between our method and 4DGS. As the scene scale increases, the per-viewpoint rendering cost of 4DGS exhibits a rapid overall growth trend. Since there are few dynamic objects in the middle segment of the test scene, the per-viewpoint reconstruction cost of 4DGF shows a slight decrease at a scale of 800 frames. In contrast, our method maintains a consistently low and stable cost. Additionally, as the number of Gaussians increases with training iterations, our S3R-GS also sustains a consistently low cost.

\paragraph{Influence of 2D Decomposition.}
To evaluate the impact of dynamic and static element decomposition using 2D boxes and NeuralODE on reconstruction quality, we evaluate our method on the novel views synthesis benchmark of KITTI dataset. We first examine a base setting that naively utilizes tracked 2D boxes to separate dynamic objects from the scene, assigning each object a learnable pose matrix at each time step. As shown in \cref{tab:abl_2dbox}, this approach struggles to accurately capture object trajectories, leading to degraded reconstruction quality. By introducing NeuralODE to model the continuous motion of objects, our method achieves competitive reconstruction quality with precise 3D bounding box separation. Compared to the complex task of acquiring 3D bounding boxes, the availability of 2D boxes significantly reduces manual annotation efforts. The qualitative comparisons are shown in the supplementary. 

\begin{table}
\renewcommand{\arraystretch}{1.05}
  \centering
  \resizebox{0.95\linewidth}{!}{
  \begin{tabular}{l|ccc}
    \hline \hline
    \rowcolor[gray]{.92} & \multicolumn{3}{c}{KITTI [75\%]} \\
     \rowcolor[gray]{.92} Method & PSNR $\uparrow$ & SSIM $\uparrow$ & LPIPS $\downarrow$ \\
    \midrule
    3D box & 31.49 & 0.946 & 0.024 \\
    \hline
    2D box & 21.09 & 0.853 & 0.111 \\
    \rowcolor{tabblue!10} 2D box + NeuralODE & 30.48 & 0.941 & 0.028\\
    \hline \hline
  \end{tabular}
  }
  \vspace{-0.2cm}
  \caption{\textbf{Ablation Study on Different Decomposition Methods.} The 2D decomposition method achieves competitive reconstruction quality compared to precise 3D bounding box separation by incorporating the NeuralODE model.}
  \label{tab:abl_2dbox}
  \vspace{-0.1cm}
\end{table}

\paragraph{Influence of BEV-semantic initialization augmentation.}
\cref{tab:abl_bev} presents the performance of our S3R-GS framework with BEV-semantic initialization augmentation on the KITTI benchmark. The incorporation of BEV-semantic initialization to supplement the initial Gaussians of tall structures in street scenes enhances the reconstruction quality, particularly for urban scenes characterized by a high density of buildings. For instance, in the KITTI 0011 scene, the introduction of BEV-semantic initialization augmentation resulted in an improvement of 0.45 dB in PSNR. 
\begin{table}
  \centering
  \renewcommand{\arraystretch}{1.05}
  \resizebox{0.95\linewidth}{!}{
  \begin{tabular}{c|ccc|ccc}
    \hline \hline
    \rowcolor[gray]{.92} & \multicolumn{3}{|c}{KITTI [75\%]} & \multicolumn{3}{|c}{KITTI 0011 [75\%]} \\
     \rowcolor[gray]{.92} Method & PSNR $\uparrow$ & SSIM $\uparrow$ & LPIPS $\downarrow$ & PSNR $\uparrow$ & SSIM $\uparrow$ & LPIPS $\downarrow$ \\
    \midrule
   w/o BEV aug. & 31.49 & \textbf{0.946} & 0.024 & 24.21 & 0.815 & 0.184 \\
    \rowcolor{tabblue!10} w BEV aug. & \textbf{31.54} & \textbf{0.946} & \textbf{0.023} & \textbf{24.66} & \textbf{0.824} & \textbf{0.171} \\
    \hline \hline
  \end{tabular}
  }
  \vspace{-0.2cm}
  \caption{\textbf{Ablation Study on the BEV-Semantic Initialization Augmentation Method.} Supplementing points in tall structure regions contributes to improved reconstruction quality.}
  \label{tab:abl_bev}
  \vspace{-0.3cm}
\end{table}

\section{Conclusion}
In this work, we propose a 3DGS framework that streamlines the pipeline for large-scale dynamic street scene reconstruction. In the scene modeling stage, we propose a BEV-semantic initialization augmentation method to accelerate reconstruction convergence. Furthermore, we utilized more accessible 2D boxes with NeuralODE model to decompose and model the scenes. In the scene reconstruction stage, we introduce the instance-specific projection, temporal visibility, and an adaptive LOD strategy to eliminate the computational redundancy. The results demonstrate that our S3R-GS outperformed existing 3DGS-based approaches in both reconstruction quality and training speed on publicly available street scene datasets.

{
    \small
    \bibliographystyle{ieeenat_fullname}
    \bibliography{main}

\begin{thebibliography}{57}
\providecommand{\natexlab}[1]{#1}
\providecommand{\url}[1]{\texttt{#1}}
\expandafter\ifx\csname urlstyle\endcsname\relax
  \providecommand{\doi}[1]{doi: #1}\else
  \providecommand{\doi}{doi: \begingroup \urlstyle{rm}\Url}\fi

\bibitem[Aliev et~al.(2020)Aliev, Sevastopolsky, Kolos, Ulyanov, and Lempitsky]{aliev2020neural}
Kara-Ali Aliev, Artem Sevastopolsky, Maria Kolos, Dmitry Ulyanov, and Victor Lempitsky.
\newblock Neural point-based graphics.
\newblock In \emph{Computer Vision--ECCV 2020: 16th European Conference, Glasgow, UK, August 23--28, 2020, Proceedings, Part XXII 16}, pages 696--712. Springer, 2020.

\bibitem[Barron et~al.(2021)Barron, Mildenhall, Tancik, Hedman, Martin-Brualla, and Srinivasan]{barron2021mip}
Jonathan~T Barron, Ben Mildenhall, Matthew Tancik, Peter Hedman, Ricardo Martin-Brualla, and Pratul~P Srinivasan.
\newblock Mip-nerf: A multiscale representation for anti-aliasing neural radiance fields.
\newblock In \emph{Proceedings of the IEEE/CVF international conference on computer vision}, pages 5855--5864, 2021.

\bibitem[Barron et~al.(2022)Barron, Mildenhall, Verbin, Srinivasan, and Hedman]{barron2022mip}
Jonathan~T Barron, Ben Mildenhall, Dor Verbin, Pratul~P Srinivasan, and Peter Hedman.
\newblock Mip-nerf 360: Unbounded anti-aliased neural radiance fields.
\newblock In \emph{Proceedings of the IEEE/CVF conference on computer vision and pattern recognition}, pages 5470--5479, 2022.

\bibitem[Barron et~al.(2023)Barron, Mildenhall, Verbin, Srinivasan, and Hedman]{barron2023zip}
Jonathan~T Barron, Ben Mildenhall, Dor Verbin, Pratul~P Srinivasan, and Peter Hedman.
\newblock Zip-nerf: Anti-aliased grid-based neural radiance fields.
\newblock In \emph{Proceedings of the IEEE/CVF International Conference on Computer Vision}, pages 19697--19705, 2023.

\bibitem[Caesar et~al.(2020)Caesar, Bankiti, Lang, Vora, Liong, Xu, Krishnan, Pan, Baldan, and Beijbom]{caesar2020nuscenes}
Holger Caesar, Varun Bankiti, Alex~H Lang, Sourabh Vora, Venice~Erin Liong, Qiang Xu, Anush Krishnan, Yu Pan, Giancarlo Baldan, and Oscar Beijbom.
\newblock nuscenes: A multimodal dataset for autonomous driving.
\newblock In \emph{Proceedings of the IEEE/CVF conference on computer vision and pattern recognition}, pages 11621--11631, 2020.

\bibitem[Chen et~al.(2018)Chen, Rubanova, Bettencourt, and Duvenaud]{neuralode}
Ricky~TQ Chen, Yulia Rubanova, Jesse Bettencourt, and David~K Duvenaud.
\newblock Neural ordinary differential equations.
\newblock \emph{Advances in neural information processing systems}, 31, 2018.

\bibitem[Cheng et~al.()Cheng, Long, Yin, Wang, Wu, Ma, Wang, Chen, and Chen]{chenguc}
Kai Cheng, Xiaoxiao Long, Wei Yin, Jin Wang, Zhiqiang Wu, Yuexin Ma, Kaixuan Wang, Xiaozhi Chen, and Xuejin Chen.
\newblock Uc-nerf: Neural radiance field for under-calibrated multi-view cameras in autonomous driving.
\newblock In \emph{The Twelfth International Conference on Learning Representations}.

\bibitem[Cornelis et~al.(2008)Cornelis, Leibe, Cornelis, and Van~Gool]{cornelis20083d}
Nico Cornelis, Bastian Leibe, Kurt Cornelis, and Luc Van~Gool.
\newblock 3d urban scene modeling integrating recognition and reconstruction.
\newblock \emph{International Journal of Computer Vision}, 78:\penalty0 121--141, 2008.

\bibitem[Dai et~al.(2020)Dai, Zhang, Li, Liu, and Zeng]{dai2020neural}
Peng Dai, Yinda Zhang, Zhuwen Li, Shuaicheng Liu, and Bing Zeng.
\newblock Neural point cloud rendering via multi-plane projection.
\newblock In \emph{Proceedings of the IEEE/CVF Conference on Computer Vision and Pattern Recognition}, pages 7830--7839, 2020.

\bibitem[Dosovitskiy et~al.(2017)Dosovitskiy, Ros, Codevilla, Lopez, and Koltun]{dosovitskiy2017carla}
Alexey Dosovitskiy, German Ros, Felipe Codevilla, Antonio Lopez, and Vladlen Koltun.
\newblock Carla: An open urban driving simulator.
\newblock In \emph{Conference on robot learning}, pages 1--16. PMLR, 2017.

\bibitem[Fischer et~al.(2024{\natexlab{a}})Fischer, Kulhanek, Bul{\`o}, Porzi, Pollefeys, and Kontschieder]{4dgf}
Tobias Fischer, Jonas Kulhanek, Samuel~Rota Bul{\`o}, Lorenzo Porzi, Marc Pollefeys, and Peter Kontschieder.
\newblock Dynamic 3d gaussian fields for urban areas.
\newblock In \emph{The Thirty-eighth Annual Conference on Neural Information Processing Systems}, 2024{\natexlab{a}}.

\bibitem[Fischer et~al.(2024{\natexlab{b}})Fischer, Porzi, Bulo, Pollefeys, and Kontschieder]{fischer2024multi}
Tobias Fischer, Lorenzo Porzi, Samuel~Rota Bulo, Marc Pollefeys, and Peter Kontschieder.
\newblock Multi-level neural scene graphs for dynamic urban environments.
\newblock In \emph{Proceedings of the IEEE/CVF Conference on Computer Vision and Pattern Recognition}, pages 21125--21135, 2024{\natexlab{b}}.

\bibitem[Fridovich-Keil et~al.(2022)Fridovich-Keil, Yu, Tancik, Chen, Recht, and Kanazawa]{fridovich2022plenoxels}
Sara Fridovich-Keil, Alex Yu, Matthew Tancik, Qinhong Chen, Benjamin Recht, and Angjoo Kanazawa.
\newblock Plenoxels: Radiance fields without neural networks.
\newblock In \emph{Proceedings of the IEEE/CVF conference on computer vision and pattern recognition}, pages 5501--5510, 2022.

\bibitem[Furukawa and Ponce(2009)]{furukawa2009accurate}
Yasutaka Furukawa and Jean Ponce.
\newblock Accurate, dense, and robust multiview stereopsis.
\newblock \emph{IEEE transactions on pattern analysis and machine intelligence}, 32\penalty0 (8):\penalty0 1362--1376, 2009.

\bibitem[Gallup et~al.(2010)Gallup, Pollefeys, and Frahm]{gallup20103d}
David Gallup, Marc Pollefeys, and Jan-Michael Frahm.
\newblock 3d reconstruction using an n-layer heightmap.
\newblock In \emph{Joint Pattern Recognition Symposium}, pages 1--10. Springer, 2010.

\bibitem[Garbin et~al.(2021)Garbin, Kowalski, Johnson, Shotton, and Valentin]{garbin2021fastnerf}
Stephan~J Garbin, Marek Kowalski, Matthew Johnson, Jamie Shotton, and Julien Valentin.
\newblock Fastnerf: High-fidelity neural rendering at 200fps.
\newblock In \emph{Proceedings of the IEEE/CVF international conference on computer vision}, pages 14346--14355, 2021.

\bibitem[Geiger et~al.(2013)Geiger, Lenz, Stiller, and Urtasun]{Geiger2013IJRR}
Andreas Geiger, Philip Lenz, Christoph Stiller, and Raquel Urtasun.
\newblock Vision meets robotics: The kitti dataset.
\newblock \emph{International Journal of Robotics Research}, 2013.

\bibitem[Guo et~al.(2023)Guo, Deng, Li, Bai, Shi, Wang, Ding, Wang, and Li]{guo2023streetsurf}
Jianfei Guo, Nianchen Deng, Xinyang Li, Yeqi Bai, Botian Shi, Chiyu Wang, Chenjing Ding, Dongliang Wang, and Yikang Li.
\newblock Streetsurf: Extending multi-view implicit surface reconstruction to street views.
\newblock \emph{arXiv preprint arXiv:2306.04988}, 2023.

\bibitem[Huang et~al.(2024)Huang, Wei, Zheng, An, Lu, Zhan, Tomizuka, Keutzer, and Zhang]{huang2024sgaussian}
Nan Huang, Xiaobao Wei, Wenzhao Zheng, Pengju An, Ming Lu, Wei Zhan, Masayoshi Tomizuka, Kurt Keutzer, and Shanghang Zhang.
\newblock Sgaussian: Self-supervised street gaussians for autonomous driving.
\newblock \emph{arXiv preprint arXiv:2405.20323}, 2024.

\bibitem[Irshad et~al.(2023)Irshad, Zakharov, Liu, Guizilini, Kollar, Gaidon, Kira, and Ambrus]{irshad2023neo}
Muhammad~Zubair Irshad, Sergey Zakharov, Katherine Liu, Vitor Guizilini, Thomas Kollar, Adrien Gaidon, Zsolt Kira, and Rares Ambrus.
\newblock Neo 360: Neural fields for sparse view synthesis of outdoor scenes.
\newblock In \emph{Proceedings of the IEEE/CVF International Conference on Computer Vision}, pages 9187--9198, 2023.

\bibitem[Kerbl et~al.(2023)Kerbl, Kopanas, Leimk{\"u}hler, and Drettakis]{kerbl20233d}
Bernhard Kerbl, Georgios Kopanas, Thomas Leimk{\"u}hler, and George Drettakis.
\newblock 3d gaussian splatting for real-time radiance field rendering.
\newblock \emph{ACM Transactions on Graphics}, 42\penalty0 (4):\penalty0 1--14, 2023.

\bibitem[Kirillov et~al.(2023)Kirillov, Mintun, Ravi, Mao, Rolland, Gustafson, Xiao, Whitehead, Berg, Lo, Doll{\'a}r, and Girshick]{kirillov2023segany}
Alexander Kirillov, Eric Mintun, Nikhila Ravi, Hanzi Mao, Chloe Rolland, Laura Gustafson, Tete Xiao, Spencer Whitehead, Alexander~C. Berg, Wan-Yen Lo, Piotr Doll{\'a}r, and Ross Girshick.
\newblock Segment anything.
\newblock \emph{arXiv:2304.02643}, 2023.

\bibitem[Kopanas et~al.(2021)Kopanas, Philip, Leimk{\"u}hler, and Drettakis]{kopanas2021point}
Georgios Kopanas, Julien Philip, Thomas Leimk{\"u}hler, and George Drettakis.
\newblock Point-based neural rendering with per-view optimization.
\newblock In \emph{Computer Graphics Forum}, pages 29--43. Wiley Online Library, 2021.

\bibitem[Lee et~al.(2024)Lee, Lee, Jo, Im, Seon, and Yoon]{lee2024semcity}
Jumin Lee, Sebin Lee, Changho Jo, Woobin Im, Juhyeong Seon, and Sung-Eui Yoon.
\newblock Semcity: Semantic scene generation with triplane diffusion.
\newblock In \emph{Proceedings of the IEEE/CVF Conference on Computer Vision and Pattern Recognition}, pages 28337--28347, 2024.

\bibitem[Lin et~al.(2022)Lin, Peng, Xu, Yan, Shuai, Bao, and Zhou]{lin2022efficient}
Haotong Lin, Sida Peng, Zhen Xu, Yunzhi Yan, Qing Shuai, Hujun Bao, and Xiaowei Zhou.
\newblock Efficient neural radiance fields for interactive free-viewpoint video.
\newblock In \emph{SIGGRAPH Asia 2022 Conference Papers}, pages 1--9, 2022.

\bibitem[Lin et~al.(2023)Lin, Peng, Xu, Xie, He, Bao, and Zhou]{lin2023high}
Haotong Lin, Sida Peng, Zhen Xu, Tao Xie, Xingyi He, Hujun Bao, and Xiaowei Zhou.
\newblock High-fidelity and real-time novel view synthesis for dynamic scenes.
\newblock In \emph{SIGGRAPH Asia 2023 Conference Papers}, pages 1--9, 2023.

\bibitem[Liu et~al.(2023{\natexlab{a}})Liu, Chen, Yang, Wang, Manivasagam, and Urtasun]{liu2023neural}
Jeffrey~Yunfan Liu, Yun Chen, Ze Yang, Jingkang Wang, Sivabalan Manivasagam, and Raquel Urtasun.
\newblock Neural scene rasterization for large scene rendering in real time.
\newblock In \emph{The IEEE International Conference on Computer Vision}, 2023{\natexlab{a}}.

\bibitem[Liu et~al.(2023{\natexlab{b}})Liu, Zeng, Ren, Li, Zhang, Yang, Li, Yang, Su, Zhu, et~al.]{liu2023grounding}
Shilong Liu, Zhaoyang Zeng, Tianhe Ren, Feng Li, Hao Zhang, Jie Yang, Chunyuan Li, Jianwei Yang, Hang Su, Jun Zhu, et~al.
\newblock Grounding dino: Marrying dino with grounded pre-training for open-set object detection.
\newblock \emph{arXiv preprint arXiv:2303.05499}, 2023{\natexlab{b}}.

\bibitem[Lu et~al.(2023)Lu, Xu, Chen, Li, Lin, and Jiang]{lu2023urban}
Fan Lu, Yan Xu, Guang Chen, Hongsheng Li, Kwan-Yee Lin, and Changjun Jiang.
\newblock Urban radiance field representation with deformable neural mesh primitives.
\newblock In \emph{Proceedings of the IEEE/CVF International Conference on Computer Vision}, pages 465--476, 2023.

\bibitem[Luiten et~al.(2024)Luiten, Kopanas, Leibe, and Ramanan]{luiten2024dynamic}
Jonathon Luiten, Georgios Kopanas, Bastian Leibe, and Deva Ramanan.
\newblock Dynamic 3d gaussians: Tracking by persistent dynamic view synthesis.
\newblock In \emph{2024 International Conference on 3D Vision (3DV)}, pages 800--809. IEEE, 2024.

\bibitem[Martin-Brualla et~al.(2021)Martin-Brualla, Radwan, Sajjadi, Barron, Dosovitskiy, and Duckworth]{martin2021nerf}
Ricardo Martin-Brualla, Noha Radwan, Mehdi~SM Sajjadi, Jonathan~T Barron, Alexey Dosovitskiy, and Daniel Duckworth.
\newblock Nerf in the wild: Neural radiance fields for unconstrained photo collections.
\newblock In \emph{Proceedings of the IEEE/CVF conference on computer vision and pattern recognition}, pages 7210--7219, 2021.

\bibitem[Mildenhall et~al.(2021)Mildenhall, Srinivasan, Tancik, Barron, Ramamoorthi, and Ng]{mildenhall2021nerf}
Ben Mildenhall, Pratul~P Srinivasan, Matthew Tancik, Jonathan~T Barron, Ravi Ramamoorthi, and Ren Ng.
\newblock Nerf: Representing scenes as neural radiance fields for view synthesis.
\newblock \emph{Communications of the ACM}, 65\penalty0 (1):\penalty0 99--106, 2021.

\bibitem[M{\"u}ller et~al.(2022)M{\"u}ller, Evans, Schied, and Keller]{muller2022instant}
Thomas M{\"u}ller, Alex Evans, Christoph Schied, and Alexander Keller.
\newblock Instant neural graphics primitives with a multiresolution hash encoding.
\newblock \emph{ACM transactions on graphics (TOG)}, 41\penalty0 (4):\penalty0 1--15, 2022.

\bibitem[Ost et~al.(2021)Ost, Mannan, Thuerey, Knodt, and Heide]{ost2021neural}
Julian Ost, Fahim Mannan, Nils Thuerey, Julian Knodt, and Felix Heide.
\newblock Neural scene graphs for dynamic scenes.
\newblock In \emph{Proceedings of the IEEE/CVF Conference on Computer Vision and Pattern Recognition}, pages 2856--2865, 2021.

\bibitem[Ost et~al.(2022)Ost, Laradji, Newell, Bahat, and Heide]{ost2022neural}
Julian Ost, Issam Laradji, Alejandro Newell, Yuval Bahat, and Felix Heide.
\newblock Neural point light fields.
\newblock In \emph{Proceedings of the IEEE/CVF Conference on Computer Vision and Pattern Recognition}, pages 18419--18429, 2022.

\bibitem[Park et~al.(2021)Park, Sinha, Hedman, Barron, Bouaziz, Goldman, Martin-Brualla, and Seitz]{park2021hypernerf}
Keunhong Park, Utkarsh Sinha, Peter Hedman, Jonathan~T Barron, Sofien Bouaziz, Dan~B Goldman, Ricardo Martin-Brualla, and Steven~M Seitz.
\newblock Hypernerf: A higher-dimensional representation for topologically varying neural radiance fields.
\newblock \emph{arXiv preprint arXiv:2106.13228}, 2021.

\bibitem[Rematas et~al.(2022)Rematas, Liu, Srinivasan, Barron, Tagliasacchi, Funkhouser, and Ferrari]{rematas2022urban}
Konstantinos Rematas, Andrew Liu, Pratul~P Srinivasan, Jonathan~T Barron, Andrea Tagliasacchi, Thomas Funkhouser, and Vittorio Ferrari.
\newblock Urban radiance fields.
\newblock In \emph{Proceedings of the IEEE/CVF Conference on Computer Vision and Pattern Recognition}, pages 12932--12942, 2022.

\bibitem[Shi et~al.(2024)Shi, Wang, Duan, and Guan]{shi2024language}
Jin-Chuan Shi, Miao Wang, Hao-Bin Duan, and Shao-Hua Guan.
\newblock Language embedded 3d gaussians for open-vocabulary scene understanding.
\newblock In \emph{Proceedings of the IEEE/CVF Conference on Computer Vision and Pattern Recognition}, pages 5333--5343, 2024.

\bibitem[Song et~al.(2023)Song, Chen, Li, Chen, Chen, Yuan, Xu, and Geiger]{song2023nerfplayer}
Liangchen Song, Anpei Chen, Zhong Li, Zhang Chen, Lele Chen, Junsong Yuan, Yi Xu, and Andreas Geiger.
\newblock Nerfplayer: A streamable dynamic scene representation with decomposed neural radiance fields.
\newblock \emph{IEEE Transactions on Visualization and Computer Graphics}, 29\penalty0 (5):\penalty0 2732--2742, 2023.

\bibitem[Tancik et~al.(2022)Tancik, Casser, Yan, Pradhan, Mildenhall, Srinivasan, Barron, and Kretzschmar]{tancik2022block}
Matthew Tancik, Vincent Casser, Xinchen Yan, Sabeek Pradhan, Ben Mildenhall, Pratul~P Srinivasan, Jonathan~T Barron, and Henrik Kretzschmar.
\newblock Block-nerf: Scalable large scene neural view synthesis.
\newblock In \emph{Proceedings of the IEEE/CVF Conference on Computer Vision and Pattern Recognition}, pages 8248--8258, 2022.

\bibitem[Turki et~al.(2022)Turki, Ramanan, and Satyanarayanan]{turki2022mega}
Haithem Turki, Deva Ramanan, and Mahadev Satyanarayanan.
\newblock Mega-nerf: Scalable construction of large-scale nerfs for virtual fly-throughs.
\newblock In \emph{Proceedings of the IEEE/CVF Conference on Computer Vision and Pattern Recognition}, pages 12922--12931, 2022.

\bibitem[Turki et~al.(2023)Turki, Zhang, Ferroni, and Ramanan]{turki2023suds}
Haithem Turki, Jason~Y Zhang, Francesco Ferroni, and Deva Ramanan.
\newblock Suds: Scalable urban dynamic scenes.
\newblock In \emph{Proceedings of the IEEE/CVF Conference on Computer Vision and Pattern Recognition}, pages 12375--12385, 2023.

\bibitem[Wang(2004)]{wang2004image}
Zhou Wang.
\newblock Image quality assessment: from error visibility to structural similarity.
\newblock \emph{IEEE transactions on image processing}, 13\penalty0 (4):\penalty0 600--612, 2004.

\bibitem[Wang et~al.(2022)Wang, Chen, Acuna, Kautz, and Fidler]{wang2022neural}
Zian Wang, Wenzheng Chen, David Acuna, Jan Kautz, and Sanja Fidler.
\newblock Neural light field estimation for street scenes with differentiable virtual object insertion.
\newblock In \emph{European Conference on Computer Vision}, pages 380--397. Springer, 2022.

\bibitem[Wang et~al.(2023)Wang, Shen, Gao, Huang, Munkberg, Hasselgren, Gojcic, Chen, and Fidler]{wang2023neural}
Zian Wang, Tianchang Shen, Jun Gao, Shengyu Huang, Jacob Munkberg, Jon Hasselgren, Zan Gojcic, Wenzheng Chen, and Sanja Fidler.
\newblock Neural fields meet explicit geometric representations for inverse rendering of urban scenes.
\newblock In \emph{Proceedings of the IEEE/CVF Conference on Computer Vision and Pattern Recognition}, pages 8370--8380, 2023.

\bibitem[Wilson et~al.(2023)Wilson, Qi, Agarwal, Lambert, Singh, Khandelwal, Pan, Kumar, Hartnett, Pontes, et~al.]{av2}
Benjamin Wilson, William Qi, Tanmay Agarwal, John Lambert, Jagjeet Singh, Siddhesh Khandelwal, Bowen Pan, Ratnesh Kumar, Andrew Hartnett, Jhony~Kaesemodel Pontes, et~al.
\newblock Argoverse 2: Next generation datasets for self-driving perception and forecasting.
\newblock \emph{arXiv preprint arXiv:2301.00493}, 2023.

\bibitem[Wu et~al.(2024)Wu, Yi, Fang, Xie, Zhang, Wei, Liu, Tian, and Wang]{wu20244d}
Guanjun Wu, Taoran Yi, Jiemin Fang, Lingxi Xie, Xiaopeng Zhang, Wei Wei, Wenyu Liu, Qi Tian, and Xinggang Wang.
\newblock 4d gaussian splatting for real-time dynamic scene rendering.
\newblock In \emph{Proceedings of the IEEE/CVF Conference on Computer Vision and Pattern Recognition}, pages 20310--20320, 2024.

\bibitem[Wu et~al.(2023)Wu, Liu, Luo, Zhong, Chen, Xiao, Hou, Lou, Chen, Yang, et~al.]{wu2023mars}
Zirui Wu, Tianyu Liu, Liyi Luo, Zhide Zhong, Jianteng Chen, Hongmin Xiao, Chao Hou, Haozhe Lou, Yuantao Chen, Runyi Yang, et~al.
\newblock Mars: An instance-aware, modular and realistic simulator for autonomous driving.
\newblock In \emph{CAAI International Conference on Artificial Intelligence}, pages 3--15. Springer, 2023.

\bibitem[Xie et~al.(2024)Xie, Chen, Hong, and Liu]{xie2024citydreamer}
Haozhe Xie, Zhaoxi Chen, Fangzhou Hong, and Ziwei Liu.
\newblock Citydreamer: Compositional generative model of unbounded 3d cities.
\newblock In \emph{Proceedings of the IEEE/CVF Conference on Computer Vision and Pattern Recognition}, pages 9666--9675, 2024.

\bibitem[Xu et~al.(2023)Xu, Xiangli, Peng, Pan, Zhao, Theobalt, Dai, and Lin]{xu2023grid}
Linning Xu, Yuanbo Xiangli, Sida Peng, Xingang Pan, Nanxuan Zhao, Christian Theobalt, Bo Dai, and Dahua Lin.
\newblock Grid-guided neural radiance fields for large urban scenes.
\newblock In \emph{Proceedings of the IEEE/CVF Conference on Computer Vision and Pattern Recognition}, pages 8296--8306, 2023.

\bibitem[Yan et~al.(2024)Yan, Lin, Zhou, Wang, Sun, Zhan, Lang, Zhou, and Peng]{yan2024street}
Yunzhi Yan, Haotong Lin, Chenxu Zhou, Weijie Wang, Haiyang Sun, Kun Zhan, Xianpeng Lang, Xiaowei Zhou, and Sida Peng.
\newblock Street gaussians: Modeling dynamic urban scenes with gaussian splatting.
\newblock In \emph{ECCV}, 2024.

\bibitem[Yang et~al.(2023)Yang, Ivanovic, Litany, Weng, Kim, Li, Che, Xu, Fidler, Pavone, et~al.]{yang2023emernerf}
Jiawei Yang, Boris Ivanovic, Or Litany, Xinshuo Weng, Seung~Wook Kim, Boyi Li, Tong Che, Danfei Xu, Sanja Fidler, Marco Pavone, et~al.
\newblock Emernerf: Emergent spatial-temporal scene decomposition via self-supervision.
\newblock \emph{arXiv preprint arXiv:2311.02077}, 2023.

\bibitem[Yang et~al.(2024)Yang, Gao, Zhou, Jiao, Zhang, and Jin]{yang2024deformable}
Ziyi Yang, Xinyu Gao, Wen Zhou, Shaohui Jiao, Yuqing Zhang, and Xiaogang Jin.
\newblock Deformable 3d gaussians for high-fidelity monocular dynamic scene reconstruction.
\newblock In \emph{Proceedings of the IEEE/CVF Conference on Computer Vision and Pattern Recognition}, pages 20331--20341, 2024.

\bibitem[Zhang et~al.(2018)Zhang, Isola, Efros, Shechtman, and Wang]{zhang2018unreasonable}
Richard Zhang, Phillip Isola, Alexei~A Efros, Eli Shechtman, and Oliver Wang.
\newblock The unreasonable effectiveness of deep features as a perceptual metric.
\newblock In \emph{Proceedings of the IEEE conference on computer vision and pattern recognition}, pages 586--595, 2018.

\bibitem[Zhenxing and Xu(2022)]{zhenxing2022switch}
MI Zhenxing and Dan Xu.
\newblock Switch-nerf: Learning scene decomposition with mixture of experts for large-scale neural radiance fields.
\newblock In \emph{The Eleventh International Conference on Learning Representations}, 2022.

\bibitem[Zhou et~al.(2024{\natexlab{a}})Zhou, Shao, Xu, Bai, Qiu, Liu, Wang, Geiger, and Liao]{zhou2024hugs}
Hongyu Zhou, Jiahao Shao, Lu Xu, Dongfeng Bai, Weichao Qiu, Bingbing Liu, Yue Wang, Andreas Geiger, and Yiyi Liao.
\newblock Hugs: Holistic urban 3d scene understanding via gaussian splatting.
\newblock In \emph{Proceedings of the IEEE/CVF Conference on Computer Vision and Pattern Recognition}, pages 21336--21345, 2024{\natexlab{a}}.

\bibitem[Zhou et~al.(2024{\natexlab{b}})Zhou, Lin, Shan, Wang, Sun, and Yang]{zhou2024drivinggaussian}
Xiaoyu Zhou, Zhiwei Lin, Xiaojun Shan, Yongtao Wang, Deqing Sun, and Ming-Hsuan Yang.
\newblock Drivinggaussian: Composite gaussian splatting for surrounding dynamic autonomous driving scenes.
\newblock In \emph{Proceedings of the IEEE/CVF Conference on Computer Vision and Pattern Recognition}, pages 21634--21643, 2024{\natexlab{b}}.

\end{thebibliography}
}

\end{document}